\documentclass[12pt]{iopart}
\usepackage[utf8]{inputenc}
\usepackage{amssymb, amsthm, mathtools, bbm, graphicx, fullpage, multirow, url, hyperref,soul} %
\usepackage{comment}
\usepackage{xcolor}
\usepackage{xurl}
\usepackage{hyperref}
\usepackage{amsmath}
\hypersetup{
    colorlinks=true,
    linkcolor=blue,
    filecolor=magenta,
    urlcolor=teal,
    citecolor=magenta,
    breaklinks=true 
}
\usepackage[utf8]{inputenc}
\usepackage[english]{babel}
\usepackage[dvipsnames]{xcolor}

\begin{document}
\title[PhysioNet Challenge 2025]{Detection of Chagas Disease from the ECG: The George B. Moody PhysioNet Challenge 2025
}

\author{
Matthew~A.~Reyna\textsuperscript{1},
Zuzana~Koscova\textsuperscript{1},
Jan~Pavlus\textsuperscript{1}, 
Soheil~Saghafi\textsuperscript{1}, 
James~Weigle\textsuperscript{1},
Andoni~Elola\textsuperscript{1,2},
Salman~Seyedi\textsuperscript{1},
Kiersten~Campbell\textsuperscript{1},
Qiao~Li\textsuperscript{1},
Ali~Bahrami~Rad\textsuperscript{1},
Ant\^onio~H.~Ribeiro\textsuperscript{3}, 
Antonio~Luiz~P.~Ribeiro\textsuperscript{4}, 
Reza~Sameni\textsuperscript{1,5,*},
Gari~D.~Clifford\textsuperscript{1,5,*}
}

\address{\textsuperscript{1}Department of Biomedical Informatics, Emory University, Atlanta, GA, USA}
\address{\textsuperscript{2}Department of Electronic Technology, University of the Basque Country UPV/EHU, Spain}
\address{\textsuperscript{3}Department of Information Technology, Uppsala University, Uppsala, Sweden}
\address{\textsuperscript{4}Universidade Federal de Minas Gerais, Belo Horizonte, Brazil; Telehealth Center from Hospital das Clínicas, Universidade Federal de Minas Gerais, Belo Horizonte, Brazil}
\address{\textsuperscript{5}Department of Biomedical Engineering, Georgia Institute of Technology and Emory University, Atlanta, GA, USA}
\address{\textsuperscript{*}These authors are joint senior authors.}

\ead{matthew.a.reyna@emory.edu}

\begin{abstract}
\emph{Objective:}
Chagas disease is a parasitic infection that is endemic to South America, Central America, and, more recently, the U.S., primarily transmitted by insects. Chronic Chagas disease can cause cardiovascular diseases and digestive problems. Serological testing capacities for Chagas disease are limited, but Chagas cardiomyopathy often manifests in ECGs, providing an opportunity to prioritize patients for testing and treatment.
\emph{Approach:}
The George B. Moody PhysioNet Challenge 2025 invites teams to develop algorithmic approaches for identifying Chagas disease from electrocardiograms (ECGs).
\emph{Main results:} 
This Challenge provides multiple innovations. First, we leveraged several datasets with labels from patient reports and serological testing, provided a large dataset with weak labels and smaller datasets with strong labels. Second, we augmented the data to support model robustness and generalizability to unseen data sources. Third, we applied an evaluation metric that captured the local serological testing capacity for Chagas disease to frame the machine learning problem as a triage task.
\emph{Significance:}
Over 630 participants from 111 teams submitted over 1300 entries during the Challenge, representing diverse approaches from academia and industry worldwide.
\end{abstract}

\section{Introduction}

The George B.\ Moody PhysioNet Challenges are annual competitions that support the development of open-source approaches to complex physiological and clinical problems \cite{Goldberger2000}. For the PhysioNet Challenge 2025, we invited teams to develop algorithms that use electrocardiograms (ECGs) to identify cases of Chagas disease to help prioritize potential Chagas patients for confirmatory diagnosis and treatment.

Chagas disease is a tropical parasitic disease that is caused by protozoan \emph{Trypanosoma cruzi} and primarily transmitted by triatomine bugs. It is endemic to South America, Central America and most recently in the U.S.~\cite{Beatty2025}. It affects more than 8 million people worldwide with 30,000 to 40,000 annual infections and 10,000 to 14,000 annual deaths \cite{Cucunuba2024, Nepomuceno2024}. There is no human vaccine for Chagas disease~\cite{camargo_why_2022, teixeira_time_2025}.

After an acute phase, which generally occurs in childhood, Chagas disease enters a life-long chronic phase \cite{Sabino2024,Nunes2018}. In the early stages of infection, Chagas disease has no or mild symptoms, and can be treated by specific drugs that can prevent the progression of the disease. In the later stages of infection, Chagas disease can cause cardiomyopathy, leading to heart failure, cardiac arrhythmias, and thromboembolism, and is associated with a higher risk of death. Serological testing has shown the widespread prevalence of Chagas disease in some areas, and such tests can be used for diagnosis in individual patients, but serological testing capacities are limited.

Given the individual and systemic harms of undiaganosed and untreated cases of Chagas disease, improving the detection of Chagas disease promises to improve outcomes and lower costs. In many countries, detection rates are below 10\%, or, more frequently, below 1\%, preventing patients from receiving timely and effective treatment \cite{world_chagas_day, teixeira_time_2025}. While serological testing is the gold standard for diagnosing Chagas disease, machine learning can help to prioritize patients for confirmatory testing.

For example, \cite{ghilardi_machine_2024} showed that simple sociodemographic and environmental risk factors can improve detection, and physiological signals can provide additional clues. Chagas cardiomyopathy, for example, often manifests in electrocardiograms (ECG), providing a signal for Chagas disease and informing the treatment of the resulting heart conditions.

Electrocardiography (ECG) provides a widely available, low-cost, and non-invasive tool to capture the electrical activity of the heart. Early research demonstrated that alterations in heart rate variability (HRV) and ECG patterns are present even before overt cardiac involvement becomes clinically evident. For example, studies have reported changes in spectral indices of HRV among Chagas patients, indicating autonomic nervous system dysfunction~\cite{4745523,Vizcardo_2019}. These early findings established HRV as a potential marker for identifying subclinical disease progression.

Approximate entropy and permutation entropy methods have been applied to Holter ECG recordings to discriminate between healthy individuals and Chagas patients, with evidence that these metrics can detect autonomic disturbances even in seropositive patients without overt ECG abnormalities~\cite{8743966,9344232,10081668, Cardoso2016}. Such approaches suggested that nonlinear characterization may enhance early risk stratification, providing insights beyond conventional diagnostic tests~\cite{Vizcardo_2019}. More recently, these entropy-based methods have been integrated with machine learning techniques to improve predictive performance~\cite{10363701,10364151}.

Parallel to these methodological advances, the broader field of artificial intelligence has demonstrated transformative potential for automated ECG analysis. Deep learning models have achieved high performance in classifying common arrhythmias and conduction abnormalities, while also enabling the prediction of latent conditions not traditionally identifiable from the ECG alone. Within the context of Chagas disease, deep learning has been applied to predict left ventricular systolic dysfunction directly from the ECG in large patient cohorts~\cite{Brito2021}, and to develop screening models capable of distinguishing seropositive individuals using only standard ECG recordings~\cite{Jidling2023}. Complementary work in animal models further supports the potential of ECG-derived markers to differentiate between acute and chronic phases of Trypanosoma cruzi infection~\cite{Haro2023}.  

These studies underscore both the promise and the challenges of leveraging ECGs for screening for Chagas disease. A particular influence on this work is the study by~\cite{Jidling2023}, which illustrates both the remarkable potential of data-driven ECG analysis to improve disease detection and the substantial difficulties posed by performance drops across independent cohorts.

For the George B. Moody PhysioNet Challenge 2025, we sought to continue efforts to use ECGs to support screening for Chagas disease.

\section{Methods}

\subsection{Data}
\label{sec:data}

For the PhysioNet Challenge 2025, we assembled 12-lead ECG recordings and Chagas disease labels from several sources. We prepared these datasets for the Challenge to create a public training set and hidden validation and test sets for the Challenge.

\subsubsection{Challenge Data Sources}
\label{sec:sources}

We used 378,624 12-lead ECG recordings recordings from 6 different sources. These sources of ECG data are described below and summarized in Table {\ref{tab:databases}. We made the training set and Chagas diagnoses for the training set publicly available, but we kept the validation and test sets hidden. The data sources in the validation and test sets have never been posted publicly, allowing us to assess common machine learning problems such as overfitting. The prevalence rates of Chagas disease in the training, validation, and test sets are approximately equal. 

The Challenge data included 12-lead ECG data and Chagas labels from multiple public and private sources:

\begin{itemize}
\item The \textbf{CODE-15\%} dataset \cite{CODE-15} contains 345779 ECG records from 233770 patients from Brazil with self-reported Chagas labels. Approximately 2\% of patients reported positive cases of Chagas disease. The signals are approximately 10s in length, and the sampling frequency is 400\,Hz. This dataset is public and part of the Challenge training set.

\item The \textbf{SaMi-Trop} dataset \cite{SaMi-Trop} contains 1631 ECG records from 1959 patients from Brazil with Chagas cardiomyopathy. All patients were serologically validated for Chagas disease. The signals are approximately 10\,s in length, and the sampling frequency is 400\,Hz. This dataset is public and part of the Challenge training set.

\item The \textbf{PTB-XL} database from the Physikalisch-Technische Bundesanstalt (PTB), Brunswick, Germany \cite{PTB-XL} contains 21799 ECG records from 18869 patients. All patients were assumed to be Chagas negative because of their geographical location in Europe, where Chagas disease is not endemic, but this assumption was not confirmed with serological testing. The signals are 10s in length, and the sampling frequency is 500\,Hz. This dataset is public and part of the Challenge training set.

\item The \textbf{REDS-II} database \cite{REDS-II} contains 1979 ECG records from 631 Brazilian patients with both positive and negative Chagas labels from serological testing. The dataset was constructed so that the numbers of Chagas positive and negative cases were approximately equal, so we oversampled the Chagas negative cases to create a positivity rate of 2\%. The signals are approximately 10\,s in length, and the sampling frequency is 300\,Hz. This dataset is private and part of the Challenge validation and test sets.

\item The \textbf{SaMi-Trop 3} database contains 3855 ECG records from Brazilian patients with both positive and negative Chagas labels from serological testing. The dataset was constructed so that the numbers of Chagas positive and negative cases were approximately equal, so we oversampled the Chagas negative cases to create a positivity rate of 2\%. The signals are approximately 10s in length, and the sampling frequency varies, including 300\,Hz, 500\,Hz, 600\,Hz, and 1000\,Hz. This dataset is private and part of the Challenge test set.

\item The \textbf{ELSA-Brasil} database \cite{ELSA-Brasil} contains 13739 ECG records from 13739 Brazilian patients with both positive and negative Chagas labels from serological testing. Approximately 2\% of patients tested positive for Chagas disease. The signals are approximately 10\,s in length, and the sampling frequency is 300\,Hz. This dataset is private and part of the Challenge test set.
\end{itemize}

\begin{table}
\centering
\begin{tabular}{|l|r|r|r|}
\hline
Database
& \multicolumn{1}{|l|}{ \begin{tabular}{@{}l@{}} Cohort \end{tabular}}
& \multicolumn{1}{|l|}{ \begin{tabular}{@{}l@{}}Recordings\end{tabular}}  & \multicolumn{1}{|l|}{ \begin{tabular}{@{}l@{}}Chagas \\ Prevalence\end{tabular}} \\\hline
CODE-15\% & Training &335,621 & 1.92 \% \\\hline
SaMi-Trop & Training & 1,631 &100.00 \%\\\hline
PTB-XL & Training &21,779 & 0.00 \%\\\hline
REDS-II &  Validation & 1,419 & 54.85 \% \\\hline
REDS-II &  Test & 560 & 50.00 \% \\\hline
SaMi-Trop 3 & Test& 3,854 & 28.50 \% \\\hline
ELSA-Brasil & Test & 13,739 & 2.04 \% \\\hline
Total &  - & 378,603 & 2.70 \%\\\hline
\end{tabular}
\caption{Summary of the Challenge data sources before preprocessing.}
\label{tab:databases}
\end{table}

\subsection{Challenge Data Preprocessing}
\label{sec:preprocessing}

For each dataset, we reformatted the data in a WFDB format so that data from different sources shared a consistent format. We truncated zero-padded ECG signals to remove added zeros, and we removed empty signals. We replaced ages above 89 with a single age of 90 as needed to deidentify the data according to the Safe Harbor method. The REDS-II dataset and the SaMi-Trop 3 dataset were constructed to artificially balance the data with comparable numbers of positive and negative Chagas cases, so we oversampled the Chagas-negative cases in these datasets to approximately match the prevalence rate of the ELSA-Brasil data, which has a 2.04\% positivity rate. For both Chagas-positive and Chagas-negative records in these two datasets, we also added small amounts of various forms of noise, applied filters that are representative of different ECG devices, and resampled the data to different sampling frequencies to create new but highly similar records. The number of recordings and the prevalence of chagas in each database after data preprocessing are shown in Table \ref{tab:databases_postprocessed}.

\begin{table}
\centering
\begin{tabular}{|l|r|r|r|}
\hline
Database 
& \multicolumn{1}{|l|}{ \begin{tabular}{@{}l@{}}Cohort\end{tabular}} & \multicolumn{1}{|l|}{\begin{tabular}{@{}l@{}}Recordings\end{tabular}} & \multicolumn{1}{|l|}{\begin{tabular}{@{}l@{}}Chagas \\ Prevalence\end{tabular}} \\\hline
CODE-15\% & Training & 335,621  & 1.92 \% \\\hline
SaMi-Trop &  Training &1,631  & 100.00 \% \\\hline
PTB-XL &  Training & 21,779  &  0.00 \% \\\hline
REDS-II  &  Validation & 37,779 & 2.05 \%\\\hline
REDS-II  &  Test & 13,720 & 2.04 \%\\\hline
SaMi-Trop 3 & Test & 53,444 & 2.06 \% \\\hline
ELSA-Brasil & Test & 13,739 & 2.04 \%\\\hline
\end{tabular}
\caption{Summary of Challenge data after preprocessing, including the data augmentation steps in Section \ref{sec:preprocessing} to preserve the Chagas disease prevalence rate in the hidden validation and test sets.}
\label{tab:databases_postprocessed}
\end{table}

\subsection{Challenge Objective}

We asked participants to design and implement working, open-source algorithms for identifying cases of Chagas disease from standard 12-lead electrocardiograms (ECG) recordings, helping to prioritize potential Chagas patients for confirmatory diagnosis and treatment.
We required teams to submit code both for their trained models and for training their models, which aided the generalizability and reproducibility of the research conducted during the Challenge. We ran the participants' trained models on the hidden validation and test sets and evaluated their performance using a tailored evaluation metric that we designed for this year's Challenge to capture the confirmatory testing capacity of areas in which Chagas disease in endemic.

\subsubsection{Challenge Overview, Rules, and Expectations}

This year's Challenge was the 26\textsuperscript{th} George B.\ Moody PhysioNet Challenge \cite{Goldberger2000}. As in previous years, the Challenge had an unofficial phase and an official phase. The unofficial phase (9 January 2025 to 9 April 2025) introduced the teams to the Challenge and provided an opportunity to discuss the topic with and seek feedback from the teams about the data, evaluation metrics, and evaluation environment. For the unofficial phase, we publicly shared the Challenge objective, training data, example algorithms, and evaluation metric and invited the teams to submit their code for evaluation on the validation set, scoring at most five entries from each team on the validation set. Between the unofficial and official phases, we took a hiatus (10 April 2025 to 28 May 2025) to improve the Challenge. The official phase (29 May 2025 to 20 August 2025) continued the Challenge and provided an opportunity for teams to refine their methods. For the official phase, we updated the Challenge data, example algorithms, and evaluation metric and again invited teams to submit their code for evaluation, scoring at most ten entries from each team on the validation set data. After the official phase, we evaluated a single entry from each team to prevent sequential training on the test data. Moreover, while teams were encouraged to ask questions, pose concerns, and discuss the Challenge in a public forum, they were prohibited from discussing their particular approaches to preserve the uniqueness of their approaches to solving the problem posed by the Challenge. 

We first ran each team's training code on the training data and then ran each team's trained code from the previous step on the hidden data. To better capture the clinical environments in which an algorithm may be deployed, we ran each algorithm sequentially on the recordings.

We allowed teams to submit either MATLAB or Python implementations of their code. Other programming languages were supported by request, but no teams requested another language. Participants containerized their code in Docker and submitted it in GitHub or Gitlab repositories. We downloaded and ran their code in containerized environments to allow teams to improve reusability of the code by allowing teams to better control their runtime environment. The computational environment is described more fully in \cite{2019ChallengeCCM}.

We used a containerized environment with 16 vCPUs, 60 GB RAM, at least 100 GB of available local storage, and an optional GPU, which was either an Ampere A30 GPU or an RTX 6000 Ada Generation GPU, depending on the available resources as the time of running the entry.

We imposed a 72-hour time limit on training with a GPU, a 96-hour time limit on training without a GPU, and a 48-hour time limit on inference for each of the validation or test sets.

To aid teams, we shared example models that we implemented in MATLAB and Python. Both the MATLAB and Python baseline model were random forest models that used the mean and standard deviation of the values in each channel of each signal as well as the available demographic information. These example models were not designed to be competitive but instead to provide examples of how to read the data and how to return the results for evaluation.

\subsubsection{The Challenge Scoring Metric}

In many Chagas-endemic regions, the availability of serological testing is severely limited due to financial, logistical, and infrastructural constraints. As a result, even if an AI system flags many patients as potentially positive based on their ECGs, only a small proportion can realistically be referred for confirmatory testing. This mismatch between diagnostic need and testing capacity fundamentally alters the design requirements for machine learning systems deployed in such settings.

To reflect this constraint, the 2025 PhysioNet Challenge evaluated algorithms based on their ability to \textit{prioritize true Chagas-positive patients within a fixed referral capacity}, rather than on traditional machine learning metrics such as the area under the receiver-operating characteristic (AUROC) curve or accuracy. Specifically, we scored each algorithm by computing the \textit{true positive rate (TPR) among the top 5\% of patients} ranked by predicted probability of Chagas disease. This 5\% corresponds to an estimated bound for the testing capacity in many real-world scenarios in Brazil: \textit{how many Chagas-positive patients can an algorithm prioritize for confirmatory testing with a constrained testing capacity?}

Formally, let $T = P + N$ denote the total number of subjects in the study, where $P$ and $N$ represent the number of positive and negative cases, respectively. Also, let $M$ denote the maximum number of subjects that can be referred for serological testing, regardless of the eventual outcome of testing. Next, suppose that each subject is assigned a risk score by a classifier, and those with scores exceeding a threshold $\tau$ are classified as positive (i.e., referred for testing). We define the \textit{true positive rate} (TPR) and the \textit{false positive rate} (FPR) at threshold $\tau$ as $\text{TPR}(\tau) = \text{TP}(\tau)/P$ and $\text{FPR}(\tau) = \text{FP}(\tau)/N$, where $\text{TP}(\tau)$ and $\text{FP}(\tau)$ are the number of true positives and false positives, respectively, at threshold $\tau$. 

To ensure the classifier operates within the referral constraint, we require that the total number of referred subjects not exceed $M$, i.e., $\text{TP}(\tau) + \text{FP}(\tau) \leq M$, or:
\begin{equation}
    \pi_p \cdot \text{TPR}(\tau) + \pi_n \cdot \text{FPR}(\tau) \leq m,
\label{eq:max_referral_chagas}
\end{equation}
where $\pi_p := P/T$ and $\pi_n := N/T$ are the class proportions for positive and negative cases, respectively, and the ratio $m := M/T$ represents the \textit{maximum admission rate}.

Equation~\eqref{eq:max_referral_chagas} defines a feasible region in ROC space bounded by a line with slope $-\pi_n/\pi_p$, intersecting the TPR axis at $m/\pi_p = M/P$ and the FPR axis at $m/\pi_n = M/N$~\cite{Sameni2025roc_geometry}, as illustrated in Figure~\ref{fig:challenge-score}. The optimal operating point of a classifier under this constraint lies at the intersection of its ROC curve with this boundary, i.e., where a given classifier detects its maximum number of true positive cases, within the testing capacity.

Rather than requiring participants to tune their threshold $\tau$ explicitly, we evaluated each algorithm's ability to identify Chagas-positive cases within the top $M$ ranked predictions. If multiple cases received the same risk score so that no threshold $\tau$ corresponded to exactly $M$ cases, then we ``broke'' the ties uniformly at random to find the expected number of positive cases.

This formulation reframes the task as a constrained ranking problem, closely reflecting deployment conditions where serological testing is scarce. The metric prioritizes high-precision triage and encourages models that are both discriminative and resource-aware---essential characteristics for scalable Chagas disease screening. As illustrated in Figure~\ref{fig:challenge-score}, the Challenge score is related to popular metrics such as the AUROC, but they are different in meaningful ways.

\begin{figure}[tb]
    \centering
    \includegraphics[width=0.4\linewidth]{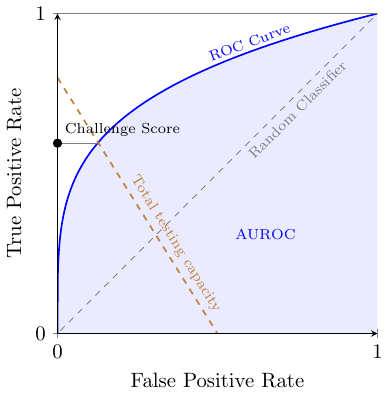}
    \caption{Illustration of the Challenge evaluation metric in receiver-operating characteristic (ROC) space. The shaded triangle represents the feasible operating region under the fixed testing capacity constraint. The Challenge score corresponds to the true positive rate (TPR) achieved within the top 5\% of predicted cases, approximating the real-world serological testing limit. Illustration adapted from~\cite{Sameni2025roc_geometry}.}
    \label{fig:challenge-score}
\end{figure}

\section{Results}

We received a total of 1317 code submissions from 111 teams with over 650 team members during the course of the Challenge. There were 451 submissions during the unofficial phase, including 185 successful submissions and 266 unsuccessful submissions with 75 submissions on the last day of the unofficial phase. There were 866 submissions during the official phase, including 372 successful submissions and 494 unsuccessful submissions with 99 submissions on the last day of the official phase.

After the end of the official phase, we attempted to score one entry from each team on the test set in Section \ref{sec:data}.
There were 65 with at least one entry that we could train on the public training set and evaluate on the hidden validation and test sets; a total of 41 teams met all of the requirements to be ranked, including an accepted CinC conference abstract, a preprint and final CinC conference proceedings paper submission, and CinC conference registration.

Tables \ref{tab:scores} summarizes the highest-ranked teams. Team summaries, additional scores, and the full Challenge criteria for rankings are available in \cite{2025ChallengeWebsite}.

\begin{table}[t]
    \centering
    \begin{tabular}{|r|l|r|r|r|r|r|l|}
        \hline
        Rank & Team name & \multicolumn{1}{|l|}{ \begin{tabular}{@{}c@{}} REDS-II \\ (validation) \end{tabular}} & \multicolumn{1}{|l|}{\begin{tabular}{@{}c@{}} REDS-II \\ (test) \end{tabular}} & \multicolumn{1}{|l|}{\begin{tabular}{@{}c@{}} SaMi-Trop 3 \\ (test) \end{tabular}} & \multicolumn{1}{|l|}{\begin{tabular}{@{}c@{}} ELSA-Brasil \\ (test) \end{tabular}} & \multicolumn{1}{|l|}{\begin{tabular}{@{}c@{}} Mean \\ test set \end{tabular}} \\\hline
        1 & Biomed-Cardio \cite{VanSantvliet2025} & 0.445 & 0.468 & 0.376 & 0.125 & 0.323 \\\hline
        2 & DlaskaLabMUI \cite{Nicolson2025} & 0.440 & 0.357 & 0.375 & 0.118 & 0.283 \\\hline
        3 & AIChagas \cite{Vazquez2025} & 0.360 & 0.382 & 0.329 & 0.129 & 0.280 \\\hline
    \end{tabular}
    \caption{Challenge scores on the validation set, which contains data from the REDS-II dataset, and test set, which contains data from the REDS-II dataset, the SaMi-Trop 3 dataset, and the ELSA-Brasil dataset, for the three highest-ranked Challenge teams.}
    \label{tab:scores}   
\end{table}

Figure \ref{fig:scores} shows the performance of each team's chosen entry on the validation set and each dataset in the test set. The median Challenge score dropped 1.4\% (from 0.279 to 0.0275) from the REDS-II data in the validation set to the REDS-II data in the test set. The median Challenge score dropped 15\% (from 0.279 to 0.236) from the REDS-II data in the validation set to the SaMi-Trop 3 data in the test set. The median Challenge score dropped 64\% (from 0.279 to 0.100) from the REDS-II data in the validation set to the ELSA-Brasil data in the test set.

\begin{figure}[tbp]
    \centering   \includegraphics{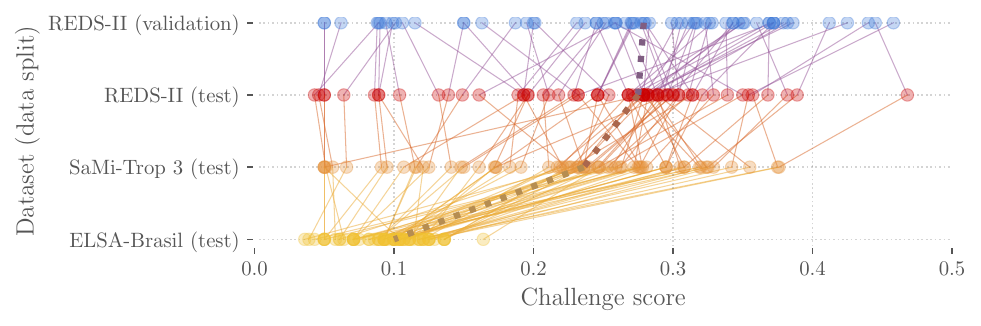}
    \caption{Challenge scores ($x$-axis) on the data sources ($y$-axis) for the hidden validation and test sets. Each point is the score of the method on a different dataset, and each thin solid line connects the scores for each team across datasets; the thick dashed color line shows the change in the median score across the different datasets.}
    \label{fig:scores}
\end{figure}

\section{Discussion}

The decreasing model performance on the test set reflects the difficulty of generalizing to unseen data. While the models perform approximately the same on the REDS-II data in the validation and test sets, they perform worse on the SaMi-Trop 3 data, which have the same prevalence rate of Chagas disease from similar patient populations with different ECG machines and collection practices, and worse still on the ELSA-Brasil data, which also have the same prevalence rate with a more asymptomatic patient population and different ECG machines and collection practices.

While model performance was generally lowest on the ELSA-Brasil data, this dataset best mirrors the potential data for a national ECG-based screening campaign for Chagas disease. Even then, the highest-performing models identified nearly three times as many Chagas-positive patients as indiscriminate testing, potentially arresting Chagas disease development in patients while they remain largely asymptomatic.

Chagas disease prevalence rates and serological testing capacities vary geographically and evolve over time. The prevalence rate of 2\% and the serological testing capacity of 5\% is roughly accurate at this time from Brazil, but interventions to reduce the transmission of Chagas disease may reduce the rates even further, increasing the benefit of such screening campaigns.

\section{Conclusions}

This article describes a large compendium of public and private 12-lead ECGs from several sources with both self-reported and serologically validated Chagas disease labels. The combination of standard 12-lead ECGs with a large database with weak labels and several smaller databases with strong labels poses a classical machine learning problem in a real-world setting, and the use of three compeletely hidden data sources with varying patient populations and environments helps to assess model generalizability to unseen data.

The use of an evaluation metric that specifically incorporates the confirmatory testing capacity for Chagas disease helps to support the development of clinical relevant machine learning models for Chagas disease detection as part of an effort to prioritize testing.

\section*{Acknowledgements}
This research is supported by the National Institute of Biomedical Imaging and Bioengineering (NIBIB: R01EB030362); the National Center for Advancing Translational Sciences of the National Institutes of Health (NCATS: UL1TR002378); as well as AliveCor and MathWorks under unrestricted gifts. AE received support by the MCIN/ AEI/10.13039/501100011033/, by FEDER Una manera de hacer Europa
through grant PID2021-122727OB-I00, and by the Basque Government through grant IT1717-22.
GDC has financial interests in AliveCor, Nextsense and Mindchild Medical and holds a board position with Mindchild Medical.
None of these entities influenced the design of or provided data for this year's Challenge.
The content of this manuscript is solely the responsibility of the authors and does not necessarily represent the official views of the above entities. 

\section*{References}
\bibliography{references} 

\end{document}